\documentclass[10pt,a4paper]{article}
\usepackage{lrec2016}
\usepackage{url}
\usepackage{latexsym}
\usepackage{graphicx}
\usepackage{arabtex}
\usepackage[T1]{fontenc}
\usepackage{enumitem}
\usepackage{colortbl,xcolor}
\usepackage{tipa}
\newcommand{\hide}[1]{}

\newcommand{\AMADDA}{{\={A}}}
\newcommand{\AHAMZAUP}{{\^{A}}}
\newcommand{\WHAMZA}{{\^{w}}}
\newcommand{\AHAMZADN}{{\v{A}}}
\newcommand{\YHAMZA}{{\^{y}}}
\newcommand{\TAMARBUTA}{{$\hbar$}}
\newcommand{\TAMAR}{{$\hbar$}}
\newcommand{\THA}{{$\theta$}}
\newcommand{\DHA}{{\dh}}
\newcommand{\SHIN}{{\v{s}}}
\newcommand{\DAD}{{\v{D}}} 
\newcommand{\AYN}{{$\varsigma$}}
\newcommand{\GAYN}{{$\gamma$}}
\newcommand{\AMAQSURA}{{\'{y}}}
\newcommand{\AMAQ}{{\'{y}}}

\newcommand{\SHADDA}{{$\sim$}}

\title{A Large Scale Corpus of Gulf Arabic}

\name{Salam Khalifa, Nizar Habash, Dana Abdulrahim$^\dagger$, Sara Hassan}

\address{Computational Approaches to Modeling Language Lab, New York University Abu Dhabi, UAE\\
	$^\dagger$University of Bahrain, Bahrain \\
         \small{\tt{\{salamkhalifa,nizar.habash,sah650\}@nyu.edu,darahim@uob.edu.bh}}\\}

\abstract{Most Arabic natural language processing   tools and resources are developed to serve Modern Standard Arabic (MSA), which is the official written language in the Arab World. Some Dialectal Arabic varieties, notably Egyptian Arabic, have received some attention lately and have a growing collection of resources that include annotated corpora and morphological analyzers and taggers. Gulf Arabic,   however, lags behind in that respect.  In this paper, we present the Gumar Corpus, a large-scale corpus of Gulf Arabic consisting of 110 million words from 1,200 forum novels. We annotate the corpus for sub-dialect information at the document level. We also present results of a preliminary study in the morphological annotation of Gulf Arabic which includes developing guidelines for a conventional orthography. The text of the corpus is publicly browsable through a web interface we developed for it. \\
\newline 
}

\begin{document}
\maketitleabstract



\setarab
\novocalize


\section{Introduction}

Most Arabic natural language processing (NLP) tools and resources are developed to serve Modern Standard Arabic (MSA),  the official written language in the Arab World. Using such tools to understand and process Dialectal Arabic (DA) is a challenging task because of the phonological and morphological differences between DA and MSA.
In addition, there is no standard orthography for DA, which only complicates matters more.
Some DA varieties, notably Egyptian Arabic, have received some attention lately and have a growing collection  of resources that include annotated corpora and morphological analyzers and taggers.  Gulf Arabic (GA), broadly defined as the variety of Arabic spoken in the countries of the Gulf Cooperation Council (Bahrain, Kuwait, Oman, Qatar, Saudi Arabia, and the United Arab Emirates), however, lags behind in that respect. 

In this paper, we present the Gumar Corpus,\footnote{Gumar \vocalize<qumar>  /\textipa{gumEr}/ is the word for `moon' in Gulf Arabic.\novocalize} a large-scale corpus of  GA that includes a number of  sub-dialects. We also present preliminary results on GA morphological annotation. Building a morphologically annotated GA corpus is a first  step towards developing NLP applications, for searching, retrieving, machine-translating, and spell-checking GA text among other applications. The importance of processing and understanding GA text (as with all DA text) is increasing due to the exponential growth of socially generated dialectal content in social media and printed works \cite{2014Sakik}, in addition to  existing materials such as folklore and local proverbs that are found scattered on the web.

The rest of this paper is structured as follows.  We present some related work in Dialectal Arabic NLP in Section~\ref{sec:related}
This is followed by a background discussion on GA in Section~\ref{sec:GAD}
We then discuss the collection of the corpus and describe its genre in Section~\ref{sec:corpus}
We present our  preliminary annotation study and evaluate it  in Section~\ref{sec:study}
Finally, we present the Gumar Corpus web interface in  Section~\ref{sec:interface}

\section{Related Work}
\label{sec:related}
\subsection{Dialectal Corpora}
 There have been many notable efforts  on the development of annotated Arabic language corpora \cite{maamouri2002resources,Maamouri:2004,PragueDep:2006,habash-roth:2009:Short,zaghouani2014large}. Most contributions however targeted MSA, developing annotation guidelines and producing large-scale Arabic Treebanks. These resources were instrumental in pushing the state-of-the-art of Arabic NLP.

Contributions that are specific to DA are limited in size, more scattered and more recent. 
Some of the earliest and relatively largest efforts have targeted  Egyptian Arabic  (EGY). 
They include CALLHOME Egyptian Arabic (CHE) corpus \cite{Gadalla1997} and its associated Egyptian Colloquial Arabic Lexicon (ECAL) \cite{ECAL:2002}.
In addition, there is the YADAC corpus \cite{Sabbagh:2012}, which was based on dialectal content identification and web harvesting of blogs, micro blogs, and forums of EGY content. And most recently, the Linguistic Data Consortium collected and annotated a sizable EGY corpus \cite{ARZMAG:2012,Maamouri:2012ARZ,ARZTB:2014}.
Levantine Arabic received less attention, with notable efforts including the 
 Levantine Arabic Treebank (LATB) of Jordanian  Arabic \cite{maamouri2006developing}
 and the Curras corpus of   Palestinian Arabic   \cite{jarrar2014building}. 
Efforts on other dialects include corpora for   Tunisian Arabic \cite{Masmoudi:2014} and Algerian Arabic \cite{smaili2014building}. 
 There are also some efforts that targeted multiple dialects such as 
 the COLABA project \cite{DiabEtAl:LREC2010-wkshp-colaba} which annotated dialectal content resources for Egyptian, Iraqi, Levantine, and Moroccan dialects from online weblogs, the Tharwa multi-dialectal lexicon \cite{diab2014tharwa},
 the multidialectal parallel Arabic corpus  
\cite{Bouamor:MDC:2014}, and the highly dialectal online commentary corpus
\cite{zaidan2011arabic}.
Most recently, in this conference proceedings, \newcite{Faisal:2016} present two morphologically annotated corpora  for Moroccan Arabic and Sanaani Yemeni Arabic.
 
As far as Gulf Arabic is concerned, 
\newcite{Halefom:2013} created an Emirati Arabic Corpus (EAC) consisting of 2 million words of transcribed Emirati TV and radio shows. The corpus was transcribed in broad IPA and translated to English. Morphological and lexical annotations as well as Arabic script annotation was manually done for a small portion of the corpus (around 15,000 words).  Furthermore, \newcite{Ntelitheos:2016} created an Emirati Arabic Language Acquisition Corpus (EMALAC) consisting of 78,000 words following the style of the widely studied CHILDES collection of corpora \cite{macwhinney2000childes}.  Both EAC and EMALAC were created by linguists with the primary purpose of studying the grammatical system of the Emirati Arabic dialect and its development. There is a lot of emphasis in the annotations of these corpora on the phonological and morphosyntactic phenomena of Emirati Arabic.  Our Gumar Corpus is differently oriented and designed: text as opposed to speech is the starting point. And computational models of GA is our target. Our corpus only includes language created by adult speakers (unlike EMALAC) and that is a slightly conventionalized novel-like form.  Finally the Gumar Corpus includes texts from a number of the Gulf countries and is not limited to the UAE.
 
For recent surveys of Arabic resources for NLP, see \newcite{zaghouani2014critical} and \newcite{shoufan2015natural}.

\subsection{Dialectal Orthography}
 Due to the lack of standardized orthography guidelines for DA, along with the phonological differences from MSA, and dialectal variations within the dialects themselves, there are many orthographic variations for written DA content. Writers in DA, regardless of the context, are often inconsistent with others and even with themselves when it comes to the written form of a dialect, writing with MSA driven orthography, or phonologically driven orthography in Arabic script or even Latin script \cite{Darwish:2013,al2014automatic}. These orthographic variations make it difficult for computational models to properly identify and reason about the words of a given dialect \cite{habash2012conventional}, hence, a conventional form for the orthographic notations is important. \newcite{habash2012conventional} proposed a Conventional Orthography for Dialectal Arabic (CODA). CODA is designed for the purpose of developing conventional computational models of Arabic dialects in general which makes it easy to be extended to other dialects. Initially, the guidelines of CODA were mainly specific to EGY. \newcite{jarrar2014building} extended the existing CODA to cover Palestinian Arabic. Recent work on  Tunisian \cite{zribi2014conventional}, Algerian \cite{saadane2015conventional} and Maghrebi Arabic \cite{MagCODA:2016}  extended the original version of CODA. We extend CODA to cover Gulf Arabic.

\subsection{Arabic Dialect Morphological Modeling}
 Most of the work that explored morphology in Arabic focused on MSA \cite{alsughaiyer/alkharashi:2004,Buckwalter:2004,SAMA31,MADAMIRA:2014}. Contributions to DA morphology analysis are usually based on either extending available MSA tools to cover DA characteristics, as in the work of \newcite{bakr2008hybrid} and \newcite{Salloum+Habash:2011}, or modeling DAs directly, without relying on existing MSA contributions \cite{habash/rambow:2006a}. One of the notable recent contributions for Egyptian Arabic morphological analysis is CALIMA \cite{Habash-CALIMA:2012}. The CALIMA analyzer for EGY and the commonly used SAMA analyzer for MSA \cite{SAMA31} are central in the functioning of the EGY morphological tagger MADA-ARZ \cite{MADA-ARZ:2013}, and its successor MADAMIRA \cite{MADAMIRA:2014}, which supports both MSA and EGY. 
\newcite{eskander2013IC} describe a technique for automatic extraction of morphological lexicons from morphologically annotated corpora and demonstrate it on EGY.   \newcite{Faisal:2016} apply the technique of \newcite{eskander2013IC} and build two morphological analyzers for Moroccan Arabic and Sanaani Yemeni Arabic.
 As for GA, we are aware of a single effort on a rule based stemmer \cite{Abuata2015stemmer} that works on sets of words collected online; they compare their results to other well known MSA stemmers.  In this paper we use MADAMIRA-EGY as a starting point for the GA morphological annotation following the approach taken by  \newcite{jarrar2014building}.


\section{Gulf Arabic Dialect}
\label{sec:GAD}
Strictly speaking, Gulf Arabic refers to the linguistic varieties spoken on the western coast of the Arabian Gulf, in Bahrain, Qatar, and the seven Emirates of the UAE \cite{qafisheh1977short}, as well as in Kuwait and in Al-Hasa -- the eastern region of Saudi Arabia \cite{holes1990gulf}. Omani, Hijazi, Najdi, and Baharna Arabic, among other additional dialects spoken in the Arabian Peninsula, are usually not included in grammars of Gulf Arabic due to the fact that they considerably vary in their linguistic features from the set of dialects listed above. In this current project, we extend the use of the term `Gulf Arabic' to include any Arabic variety spoken by the indigenous populations residing the six countries of the Gulf Cooperation Council: Bahrain, Kuwait (KW), Oman (OM), United Arab Emirates (AE), Qatar (QA), and Kingdom of Saudi Arabia (SA).  

The cultural homogeneity of the Gulf region does not necessarily entail linguistic homogeneity. Indeed, GA dialects extensively differ in their morpho-phonological and lexical features, reflecting a number of geographical and social factors \cite{holes1990gulf} in addition to being influenced by different contact languages at different time periods. A number of linguistic features set GA dialects apart from other dialects spoken in the Arab world. One of the 
distinguishing phonological features of most GA dialects includes maintaining the pharyngealized fricative /\textipa{z}\textsuperscript{\textipa{Q}}/,\footnote{Phonetic transcription is presented in IPA.}
  as well as the interdental /\textipa{T}/ and /\textipa{D}/, unlike what happens in other Arabic dialects. Among the most prominent phonological features in GA are the variant pronunciations of the sounds /\textipa{q}/, /\textipa{dZ}/, /\textipa{S}/, and /\textipa{k}/. /\textipa{q}/ may be realized in certain dialects as /\textipa{g}/ as in /\textipa{ga:l}/ `he said', or as /\textipa{dZ}/ as in /\textipa{dZIdIr}/ `pot'. /\textipa{dZ}/, on the other hand, may be realized in some varieties as /j/, as in /\textipa{jImEl}/ `camel'. The palatal /\textipa{tS}/ and the velar /\textipa{k}/ may both turn into the alveopalatal /\textipa{tS}/ as in /\textipa{tSa:j}/ `tea', and /\textipa{tSEf}/ `palm'. Moreover, The 2\textsuperscript{nd} singular feminine possessive and object pronoun /\textipa{kI}/ retains its phonological form in certain dialects (e.g. some Saudi dialects) but is realized in some other dialects as /\textipa{tS}/, /\textipa{S}/, or /\textipa{ts}/. 
 
In terms of the morphological features, and as in the case with most spoken Arabic dialects, GA dialects have also lost case inflection (with the exception of some Bedouin dialects, e.g. /\textipa{bIntIn P@s\textsuperscript{Q}i:lE}/ `respectable girl'). Possession may also be marked by clitics such as\hide{<mAl> \textit{mAl}} /\textipa{ma:l}/ and\hide{<.hg> \textit{Hg}} /\textipa{\textcrh ag}/ \cite{holes1990gulf}, e.g. /\textipa{lI-kta:b ma:lEl m2r2h}/ `The book of the lady'. Negation is also marked by the particles /\textipa{mu:}/ (and its variants /\textipa{mUb}/, /\textipa{mUhUb}/, and /\textipa{hUb}/) \cite{holes1990gulf}. The plural and the dual masculine and feminine forms of verbs and nouns are collapsed into one form in most dialects, but some distinctions are still maintained in certain others. For instance, in some varieties of Saudi, Emirati, and Omani Arabic, verbs and possessive pronouns inflected for the 3\textsuperscript{rd} plural feminine are quite distinct from the masculine forms:
 \begin{itemize}
  \item /\textipa{ga:mEt}/ `She stood up' - /\textipa{ga:mEn}/ `They[2FP] stood up'.
  \item /\textipa{wEdZhIk}/ `Your[2FS] face' - /\textipa{w@dZu:hkIn}/ `Your[2FP] faces'. 
  \end{itemize}
 Additional morphological features include the cliticized /\textipa{ba}/ for future (instead of the standard /\textipa{sa}/ prefix). Alternatively, in some varieties of GA, the motion verb /\textipa{ra:\textcrh}/ has grammaticalized into a future marker (e.g. Kuwaiti, Bahraini, and some varieties of Saudi Arabic). Less prominently, yet still a distinctive feature in some GA dialects, is the the epenthetic /\textipa{n}/ found in the active participle in some varieties of Emarati and Baharna Arabic: /\textipa{ma:Qt\textsuperscript{Q}Inh@m}/ `I've/'d given them'.
\novocalize
Finally, the lexicon of GA consists of standard Arabic cognates that may or may not follow the phonotactics of the respective dialects. Unsurprisingly, many cognate expressions that are highly frequent in a given dialect have reduced in form, such as 
\begin{itemize}[noitemsep,nolistsep]
	\item[-] /\textipa{liPay SayP}/ `For which thing'$\rightarrow$/\textipa{le:S}/ `Why'.
\end{itemize}
As for lexical borrowings in GA dialects, there is an undoubtedly substantial amount of lexical items that have been borrowed from various contact languages throughout different historical periods, e.g.,
\begin{itemize} 
 	\item /\textipa{QEmbElu:s\textsuperscript{Q}}/ `Ambulance' from English.
 	\item /\textipa{hEst}/ `There is' from Persian.
	\item /\textipa{Pa:lu:}/ `Potatoes' from Hindi.
 \end{itemize}

\section{Corpus Description}
\label{sec:corpus}
\begin{table}
\begin{center}
\begin{tabular}{|l|c|}
\hline
\multicolumn{2}{|c|}{  Gumar Corpus } \\ \hline
Words & 112,410,688 \\ \hline
Sentences & 9,335,224 \\ \hline
Documents & 1,236 \\ \hline
\end{tabular}
\caption{Statistics on the Gumar Corpus}
\label{GACN_stats}
\end{center}
\end{table}

\paragraph{Corpus Collection} Gulf Arabic, just like any other Arabic dialect has no written convention nor is it used as a formal mean of communication in the media, education or official documents. Hence there are no known go-to resources. A unique genre of written materials that is specifically known to GA is online anonymous publicly published long conversational novels. We have found a huge collection of these novels online in one place.\footnote{\url{www.graaam.com}} We automatically downloaded about 1,200  MS Word documents. Usually, such novels are written in lengthy threads that can be found in online forums. The data we got was  collected by volunteering forum members into MS Word documents and then published by another member in an organized matter.\footnote{There are no copyright claims by the anonymous writers or organizers; and we do not claim any copyrights to the text. We will make the cleaned up and extended versions of the data fully publicly available.}

\paragraph{Corpus Genre}
The main theme of most of the novels is romantic; but they  also include drama and  tragedy. The structure of a novel is simple. It starts with a brief introduction that contains the title of the novel, the writer's pen name (no real names are used) and the country of the novel. The introduction is then followed by a prologue that usually contains a small piece of dialectal poetry or a small piece of literary writing usually in MSA. It also contains a brief description of the novel characters, though some writers prefer to introduce the characters as their role appears. Then comes the main body of the novel, which is often a dialogue between the characters. There are also some pieces of narration between conversations in either the dialect or MSA. The last part of the novel usually has some "moral" lessons narrated by the writer. Writers  tend to ask the audience for positive criticism and opinions and whether they should continue writing more novels or not.

The novels are entirely written in DA except for the parts mentioned above. The dialect of the novel is not necessarily the same as the dialect of the writer. Most of the time the writers remain anonymous under nicknames, though they ask to be credited if the novel is transferred to another forum. Hence some writers are quite famous among the audience. The targeted audience is mainly female teenagers, the nature of publishing the novels is highly interactive and dependent on the activity of the audience. The writer usually ends each "part" in the novel with a teaser and demands  participation and encouragement from the audience.
Table~\ref{GACN_stats} shows  statistics on all the collected text. Words are whitespace tokenized and the counts include punctuation. The number of sentences represents the number of lines. Most of the time, each document represents a single novel; but in few cases a novel may be split into more than one document.

\paragraph{Corpus Dialects and Dialect Annotation}
We have annotated the corpus on the level of documents for the dialect, novel name and writer name for each. The dialect of the written text was the most challenging to know. In some documents, the dialect or the country of the writer was explicitly stated; in others, names of cities clearly indicated the origin country. However, in many cases,  further investigation was needed.
The GA dialects are closely intertwined, yet when thoroughly observed show evident differences. These differences were observed through common trends in relation to each GA dialect. It is important to point that the names given to the characters in each story have shown a trend with the dialect used, for example: SA dialect novels have repeatedly used the names {\small<fy.sl>}~fySl\footnote{Arabic transliteration is presented in the
  Habash-Soudi-Buckwalter scheme \cite{HSB-TRANS:2007}: (in alphabetical order)\\
  \addtolength{\tabcolsep}{-5.3pt}
\begin{tabular}{cccccccccccccccccccccccccccc}
<'a> & <b> & <t> & <_t> & <^g> & <.h> & <_h> & <d> & <_d> & <r> & <z> & <s> & <^s> & <.s> & <.d> & <.t> & <.z> & <`> & <.g> & <f> & <q> & <k> & <l> & <m> & <n> & <h> & <w> & <y> \\
{\AHAMZAUP} & b & t & {\THA} & j & H & x & d & {\DHA} & r & z & s & {\SHIN} & S & D & T & {\DAD} & {\AYN} & {\GAYN} & f & q & k & l & m & n & h & w & y\\
\end{tabular}
\\
 and the additional symbols: '~<"'>, {\AHAMZAUP}~<'a>,
  {\AHAMZADN}~<'i>, {\AMADDA}~<'A>, {\WHAMZA}~<|u'u>, {\YHAMZA}~<|Y'>,
  {\TAMARBUTA}~<T>, {\AMAQSURA}~<Y>. }
   /\textipa{fe:s\textsuperscript{Q}El}/ `Faisal', {\small<`bd Al`zyz>} {\AYN}bd Al{\AYN}zyz /\textipa{QEbdIlQEzi:z}/ `Abdulaziz' and {\small<trky>} trky /\textipa{tIrkej}/ `Turky' for male characters and {\small<lmA>} lmA /\textipa{lEmE}/ `Lama', {\small<wsn>} wsn /\textipa{wEsEn}/ `Wasan' and {\small<njwd>} njwd /\textipa{ndZu:d}/ `Njood' for female characters. On the other hand the AE dialect novels have used the names {\small<hzzA`>} hz{\SHADDA}A{\AYN} /\textipa{hEzza:Q}/ `Hazza`', {\small<zAyd>} zAyd /\textipa{za:jId}/ `Zayed' and {\small<rA^sd>} rA{\SHIN}d /\textipa{ra:SId}/ `Rashid' as male names and {\small<.h.s.sT>} HS{\SHADDA}{\TAMAR} /\textipa{\textcrh Is\textsuperscript{Q}s\textsuperscript{Q}E}/ `Hessa', {\small<my_tA'>} my{\THA}A' /\textipa{me:TE}/`Maytha' and {\small<^smmT>} {\SHIN}m{\SHADDA}{\TAMAR} /\textipa{SEmmE}/ `Shamma' for female characters. It is worth mentioning that in both SA and AE dialect novels the male names could be related to the leaders of each country and hence this may be a cultural influence that could be related to the author's intuition when selecting character names. The KW dialect novels used the names {\small<.dAry>} DAry /\textipa{d\textsuperscript{Q}a:rej}/ `Dhari', {\small<m^sAry>} m{\SHIN}Ary /\textipa{mSa:rej}/ `Mshari' and {\small<.sbA.h>} SbAH /\textipa{s\textsuperscript{Q}Uba:\textcrh}/ `Subah' for male characters while the names {\small<sbyjT>} sbyj{\TAMAR} /\textipa{sIbi:tSE}/ `Sebichah' and {\small<fwzyyT>} fwzy{\SHADDA}{\TAMAR} /\textipa{fozIjjE}/ `Fawziyya' were noticed to be prominent for female characters. 
The second noticed trend is the use of Hijri dates in SA dialect novels, which is explained by the country's official calendar use of the Hijri date. This trend was only noticed in SA dialect novels.
Thirdly, the use of commonly spoken words that could be directly traced to a particular dialect such as {\small<hAl.hzzT>} hAlHzz{\TAMAR} /\textipa{hEl\textcrh EzzE}/ `Now' and {\small<yA m`wd>} yA m{\AYN}w{\SHADDA}d /\textipa{ja: mQEwwEd}/ `O man' in KW dialect, {\small<bzrAn>} bzrAn /\textipa{bIzra:n}/ `Little kids', {\small<wrAk>} wrak /\textipa{wEra:k}/ as in {\small<wrAk mA tjAwb>} wrAk mA tjAwb /\textipa{wEra:k ma: tdZa:wIb}/ `Why are you[2MS] not answering?' which is traced to SA dialect and AE words such as {\small<sydA>} sydA /\textipa{si:dE}/ `Straight forward', {\small<Aqrb>} Aqrb /\textipa{Igr@b}/ `Come in, please!' and {\small<xA^swqT>} xA{\SHIN}wq{\TAMAR} /\textipa{xa:Su:gE}/ `Spoon'.
The above are trends noticed when annotating over a thousand GA novels that helped in adding efficiency to the task. These trends were noticed as the process progressed and selecting a dialect was only completed once parts of the story were read, alongside facts provided by the author, which include providing an insight to the readers that the story will be written in a particular dialect and/or constantly providing details of where events took place (i.e. city and/or country names).

Following on the annotation effort, we present the distribution of the dialects across the corpus, see Table~\ref{DA_stats}. We have observed that 92\% of the entire corpus is actually written in GA with SA being the most dominant and BH the least. There is also around 10\% that is identified as GA (other) which are the cases of a novel containing a combination of several GA dialects that is due to multiple writers with different dialects or due to the existence of different characters in the novel. It was sometimes hard to differentiate through the text between the three dialects of OM, QA and AE even with a native speaker annotating, hence these cases we marked as GA (other) also. The rest of the corpus (7.93\%) is mostly MSA (original text or translation attempts of existing non Arabic text) and other DA such as Egyptian, Iraqi, Levantine, ... etc.
\begin{table}
\begin{center}
\begin{tabular}{|l|c|}
\hline
Dialect & Percentage \\ \hline
SA & 60.52 \\ \hline
AE & 13.35 \\ \hline
KW & 5.91 \\ \hline
OM & 1.13 \\ \hline
QA & 0.65 \\ \hline
BH & 0.49 \\ \hline
GA (other) & 10.03 \\ \hline
Arabic (other) & 7.93 \\ \hline
\end{tabular}
\caption{Distribution of Dialects across the Gumar Corpus}
\label{DA_stats}
\end{center}
\end{table}

\section{Preliminary Investigation into GA Annotation}
\label{sec:study}
We describe next a pilot study in semi-automatic annotation of GA. 
We use the MADAMIRA \cite{MADAMIRA:2014} morphological tagger (which works in two modes: MSA and EGY)
and manually change its output in accordance to the orthography and morphology guidelines that are discussed next in this section.
\paragraph{Orthography Guidelines}
GA speakers who write in the dialect produce spontaneous inconsistent spellings that sometimes reflect the phonology of the GA, and other times the word's cognate relationship with MSA.
We follow the work of \newcite{habash2012conventional} for CODA in order to overcome these inconsistencies. There are the general CODA rules that apply for every dialect, among which is affix spelling. Affix spelling includes the spelling of the Ta Marbuta; if the Ta Marbuta is at the end of the word it will always be <T>  {\TAMAR} and not <h> h regardless of the pronunciation. When inside a word (before a clitic) Ta Marbuta will become <t-->, (e.g. {\small<syyArT .h.s.sT jdydT>} syAr{\TAMAR} HS{\TAMAR} jdyd{\TAMAR} /\textipa{sEjja:rEt \textcrh is\textsuperscript{Q}s\textsuperscript{Q}ah jIdi:dEh}/ `Hissa's car is new', {\small<syyArthA>} syArthA /\textipa{sajja:retha:}/ `her car'). \hide{ Another common inconsistency is the hamza spelling that is in the beginning of a word, we follow a later discussion on the Palestinian Arabic CODA \cite{Habash:2015:PALCODA} which states that most of the time the absence of a hamza will not create ambiguity (e.g. {\small<'ax_d>} {\AHAMZAUP}x{\DHA} vs {\small<Ax_d>} Ax{\DHA} /\textipa{PExED}/ `he took'). However, hamzas that are part of the word root are kept such as {\small<s'al>} s{\AHAMZAUP}l /\textipa{sEPEl}/ `He asked'.}  
Following on the discussion about GA dialect properties in section \ref{sec:GAD}, we extend several aspects of the original CODA. The root consonant mapping rules are extended to cover the GA pronunciations that are unseen in other dialects, see Table~\ref{GA_CODA}.
Another aspect is the spelling of the 2nd person singular feminine pronominal clitic; if spelled differently than the <ki> k /\textipa{kI}/
equivalent in MSA, it is mapped to <j> j /\textipa{dZ}/, (e.g. {\small<ktAb^s>} ktAb{\SHIN} /\textipa{kta:bIS}/, {\small<ktAbts>} ktAbts /\textipa{kta:bIts}/, and {\small<ktAbj>} ktAbj /\textipa{kta:bItS}/ `Your[FS] book', becomes {\small<ktAbj>} ktAbj /\textipa{kta:bItS}/ but not {\small<ktAbk>} ktAbik /\textipa{kta:bIk}/) in CODA.
%
As with the original CODA for EGY, we also maintain a list of exceptional spellings for uniquely dialectal words. One example in GA  is the spelling of the perfective verb {\small<kAn>} kAn /ka:n/ `was' and its  variant {\small<jAn>} jAn /\textipa{tSa:n}/: the perfective verb form is used as in MSA, however the other variant is considered a modal auxiliary \cite{Brustad:2000}. Both spellings are kept due to the difference in their usage despite the fact that they   share the same origin. Another 
example is the the negation particles {\small<mb>} mb /\textipa{mUb}/ `not' and {\small<mAnyb>} mAnyb /\textipa{mEni:b}/ `I'm not', both of which have a number of non-CODA variants such as {\small<mwb>} mwb  and {\small<mnyb>} mnyb, respectively.  The complete guidelines for the GA CODA will be available separately as a technical report. An example of the application of several CODA rules is presented in Table~\ref{CODA_example} 

\begin{table}
\setlength{\tabcolsep}{2pt}
\begin{center}
\begin{tabular}{|l|c|c|}
\hline
\bf CODA & \bf Pron. variation & \bf Example \\ \hline
<q> {\it q} & /\textipa{q}/ or /\textipa{g}/ or /\textipa{dZ}/ & <qdAm> \textit {qid{\SHADDA}Am} /\textipa{dZIdda:m}/ `Front'\\ 
 & &  <qlm> \textit {elm} /\textipa{gEl\textsuperscript{Q}Em}/ `pen(cil)' \\ \hline
<k> {\it k} & /\textipa{k}/ or /\textipa{tS}/ or /\textipa{ts}/ & <kbd> \textit {kbd} /\textipa{tSEbd}/ `Liver' \\ \hline
<j> {\it j} & /\textipa{dZ}/ or /\textipa{j}/ & <jls> \textit{jls} /\textipa{jIlas}/ `He sat' \\ \hline
<^s> {\it {\SHIN}} & /\textipa{S}/ or /\textipa{tS}/ & <^sAy> \textit{{\SHIN}Ay} /\textipa{tSa:j}/ `Tea' \\ \hline
\end{tabular}
\end{center}
\caption{GA root consonants mapping rules}
\label{GA_CODA}
\end{table}
\begin{table*}[t]
\setlength{\tabcolsep}{2pt}
\begin{center}
\begin{footnotesize}
\begin{tabular}{|l|c|c|c|}
\hline
Example 1 & Raw &\ldots <Asm` hAl.htsy mnts yAwylts> \ldots\\
 & & \ldots Asm{\AYN} hAlHtsy mnts yAwylts \ldots\\ \hline
 & CODA & \ldots <Asm` hAl.hky mn^g yAwyl^g> \ldots\\
 & & \ldots Asm{\AYN} hAlHky mnj yAwylj \ldots\\ \hline 
 & English & [If] I hear this talk from you [again][,] you will suffer \\ \hline \hline
 Example 2 & Raw &<`sY Al.gdY ^gAhz?> \ldots \\
 & & \ldots {\AYN}s{\AMAQ} Al{\GAYN}d{\AMAQ} jAhz? \\ \hline
 & CODA & <`sY Al.gdA ^gAhz?> \ldots \\
 & & \ldots {\AYN}s{\AMAQ} Al{\GAYN}dA jAhz? \\ \hline
  & English & \ldots Is lunch ready? \\ \hline \hline
 Example 3 & Raw & <sArh: mnyb .s.gyr|rwnh AnA Al|l.hyn fy Al^gAm`h> \\
 & & sArh: mnyb S{\GAYN}yrrwnh AnA AllHyn fy AljAm{\AYN}h \\ \hline
 & CODA & <sArT: mAnyb .s.gyrwnT AnA Al.hyn fy Al^gAm`T> \\
 & & sAr{\TAMAR}: mAnyb S{\GAYN}yrwn{\TAMAR} AnA AlHyn fy AljAm{\AYN}{\TAMAR} \\ \hline
 & English & Sarah: I'm not a child I'm now in university. \\ \hline

\end{tabular}
\caption{Example of application of CODA rules}
\label{CODA_example}
\end{footnotesize}
\end{center}
\end{table*}

\paragraph{Morphology Guidelines}
 For every input word MADAMIRA produces a list of analyses specifying every possible morphological interpretation of that word, covering all morphological features of the word (diacritization, part-of-speech (POS), lemma, and 13 inflectional and clitic features). MADAMIRA then applies a set of models (support vector machines and N-gram language models) to produce a prediction, per word in-context, for different morphological features, such as POS, lemma, gender, number or person. A ranking component scores the analyses produced by the morphological analyzer using a tuned weighted sum of matches with the predicted features. The top-scoring analysis is chosen as the predicted interpretation for that word in context \cite{MADAMIRA:2014}.
%
%

We follow a similar approach that was used to morphologically annotate both the EGY and LEV corpora. We select the following set of features with their initial values from the output of MADAMIRA-EGY on a given GA text to annotate: \hide{word orthography, morpheme tokenization, part-of-speech, lemma and English gloss.}
 
\begin{itemize}
\item {\bf Word orthography} We follow the previously discussed orthography guidelines. \hide{(starting point is the [DIAC] feature without diacritization)}
\item \hide{ATB tokenization +}{\bf Morphemic tokenization} A word is split into its morphemes and stem. \hide{(Extracted from the Buckwalter tag [BW])}
\item {\bf Part of Speech} We use the MADAMIRA POS tag set \hide{[POS]}
\item {\bf CATiB 6 POS} Except the tag for passive verbs  \cite{habash-roth:2009:Short}.
\item {\bf Lemma} Diacritized form of the lemma. \hide{[LEX]}
\item {\bf English gloss} The English translation of the lemma \hide{[GLOSS]}
\end{itemize}
Beside the above guidelines, there exist cases of erroneous merging and splitting of words that gets no analysis from the automatic annotation. For merged words, we place a `\#' symbol where a split should happen, this fix aggregate through all the other annotated features. In the case where there is a split, we place the `\#' symbol at the end of the first split part to indicate the merging position. 

Table~\ref{exmp_Anno} shows an annotation example following the above guidelines. Where green and red shaded cells indicates a change. 
\paragraph{Evaluation}
%

We conduct an evaluation of the quality of automatic morphological annotation tools (taggers) on this corpus to assess the amount of effort needed to manually annotate it.
Following the annotation guidelines discussed above, we  manually annotated around 4K words from four different novels with a goal to capture different dialects, styles of writing, \ldots etc.
An example of an annotated sentence is shown in Table~\ref{exmp_Anno}. 
As a preliminary experiment, we investigated the frequency of out-of-vocabulary (OOV) words in both systems of MADAMIRA: MSA and EGY. 
The Egyptian model OOV (5.6\%) was almost half that of MSA (9.3\%), suggesting it is better to work with Egyptian as the base system for manual annotation.

\begin{table*}
\begin{center}
\begin{scriptsize}
\begin{tabular}{|l|l|l|l|l|l|l|l|}
\hline
\multicolumn{8}{|c|}{\bf MADAMIRA-EGY}\\ \hline
\multicolumn{2}{|l|}{\bf Raw} & \bf CODA & \bf Morph & \bf POS & \bf CATiB 6& \bf Lemma & \bf Gloss \\ \hline
<zyAd> & zyAd & zyAd & zyAd & noun\_prop & PROP & ziyAd & Ziad\\ \hline
: & : & : & : & punc & PNX & : & :\\ \hline
<lysA'> &lysA' & lysA' & lysA' & \cellcolor{red!25}noun & \cellcolor{red!25}NOM & \cellcolor{red!25}{\AHAMZAUP}aloyas & \cellcolor{red!25}valiant\\ \hline
<mysA'> & mysA' & \cellcolor{red!25}mAysA' & \cellcolor{red!25}mA+y+sA' & \cellcolor{red!25}verb & \cellcolor{red!25}VRB & \cellcolor{red!25}{\AHAMZAUP}asA' & \cellcolor{red!25}be\_harmed\\ \hline
<AntwlAzm> & AntwlAzm & \cellcolor{red!25}NOAN & \cellcolor{red!25}NOAN & \cellcolor{red!25}NOAN & \cellcolor{red!25}NOAN & \cellcolor{red!25}NOAN & \cellcolor{red!25}NOAN\\ \hline
<tjAbwn> & tjAbwn & \cellcolor{red!25}tjAbwn & \cellcolor{red!25}t+jAb+wn & verb & VRB & \cellcolor{red!25}{\AHAMZAUP}ajAb & \cellcolor{red!25}be\_answered\\ \hline
<`> & {\AYN} & {\AYN}lY & {\AYN}lY &prep & PRT & {\AYN}alaY & on\\ \hline
<kl> & kl & kl & kl & noun\_quant & NOM & kul{\SHADDA} & all\\ \hline
<Aly> & Aly & Ally & Ally & pron\_rel & NOM & All{\SHADDA}iy & which\\ \hline
<yqwlkm> & yqwlkm & \cellcolor{red!25}yqwlkm & \cellcolor{red!25}y+qwl+km & \cellcolor{red!25}verb & \cellcolor{red!25}VRB & \cellcolor{red!25}qAl & \cellcolor{red!25}said\\ \hline
\multicolumn{8}{c}{ }\\ \hline
\multicolumn{8}{|c|}{\bf Manual Annotation}\\ \hline 
\multicolumn{2}{|l|}{\bf Raw}  & \bf CODA & \bf Morph & \bf POS & \bf CATiB 6& \bf Lemma & \bf Gloss \\ \hline
<zyAd> & zyAd & zyAd & zyAd & noun\_prop & PROP & ziyAd & Ziad\\ \hline
: & : & : & : & punc & PNX & : & :\\ \hline
<lysA'>&  lysA' & lysA' & lysA' & \cellcolor{green!25}noun\_prop & \cellcolor{green!25}PROP & \cellcolor{green!25}laysaA' & \cellcolor{green!25}Laysaa\\ \hline
<mysA'> & mysA' & \cellcolor{green!25}mysA' & \cellcolor{green!25}mysA' & \cellcolor{green!25}noun\_prop & \cellcolor{green!25}PROP & \cellcolor{green!25}MayosaA' & \cellcolor{green!25}Maysaa\\ \hline
<AntwlAzm> & AntwlAzm & \cellcolor{green!25}Antw\#lAzm & \cellcolor{green!25}Antw\#lAzm & \cellcolor{green!25}pron\#noun & \cellcolor{green!25}NOM\#NOM & \cellcolor{green!25}Antw\#lAzim & \cellcolor{green!25}you\#necessary\\ \hline
<tjAbwn> & tjAbwn & \cellcolor{green!25}tjAwbwn & \cellcolor{green!25}t+jAwb+wn & verb & VRB & \cellcolor{green!25}jAwab & \cellcolor{green!25}comply\\ \hline
<`> & {\AYN} & {\AYN}lY & {\AYN}lY &prep & PRT & {\AYN}alaY & on\\ \hline
<kl> & kl & kl & kl & noun\_quant & NOM & kul{\SHADDA} & all\\ \hline
<Aly> & Aly & Ally & Ally & pron\_rel & NOM & All{\SHADDA}iy & which\\ \hline
<yqwlkm> & yqwlkm & \cellcolor{green!25}yqwlh\#lkm & \cellcolor{green!25}y+qwl+h\#l+km & \cellcolor{green!25}verb\#prep & \cellcolor{green!25}VRB\#PRT & \cellcolor{green!25}qAl\#la & \cellcolor{green!25}said\#to\\ \hline 

\end{tabular}
\end{scriptsize}
\end{center}
\caption{Example of manual annotation following the orthography and morphology guidelines. Columns represent features to be annotated and rows represent words. NOAN means that  no analysis was given automatically.}
\label{exmp_Anno}
\end{table*}

\begin{table}[h]
\setlength{\tabcolsep}{2pt}
\begin{center}
\begin{tabular}{|l|c|c|}
\hline
\bf Feature & \bf MADAMIRA-MSA & \bf MADAMIRA-EGY\\ \hline\hline
CODA & 83.81 & 88.34\\ \hline
Morph & 76.16& 83.62\\ \hline
POS & 72.37 & 80.39 \\ \hline
CATiB & 76.28 & 81.51\\ \hline
Lemma & 64.03 & 77.02 \\ \hline
\end{tabular}
\end{center}
\caption{Results on evaluation of our Gold annotation against the output of MADAMIRA in both modes: MSA and EGY}
\label{Evaluation}
\end{table}

 We evaluated using the accuracy measure for word orthography, morphemic tokenization, POS, CATiB POS and lemma against the output of both models of MADAMIRA: MSA and EGY. Table~\ref{Evaluation} shows the results of the the evaluation for all words. 
%
These numbers allow us to assess the basic quality of the these tools on GA.  As expected, MADAMIRA-EGY outperforms MADAMIRA-MSA between 4 and 13\% absolute on different metrics, confirming that it is better to use it as a baseline. This is similar to results reported by \newcite{jarrar2014building} on Palestinian Arabic.

\paragraph{Error Analysis}
We manually investigated the four sets of 100 words from different parts of the MADAMIRA-EGY annotated sub-corpus (total 400 words, with an average of 30 words containing at least one error).
52.1\% of the errors are likely due the wrong assignment of POS especially in proper nouns that look like nouns or adjectives. Another major source of error is the lemma 18.2\% and out of vocabulary related errors are 16.5\% and this happens for two main reasons, either the word is never seen before or because of a typo. Finally, errors that come from a mistake in merging or splitting word tokens, typos and tokenization words combine around 13\%.

\section{Gumar Interface}
\label{sec:interface}
\begin{figure*}
\begin{center}
\includegraphics[width=0.84\textwidth]{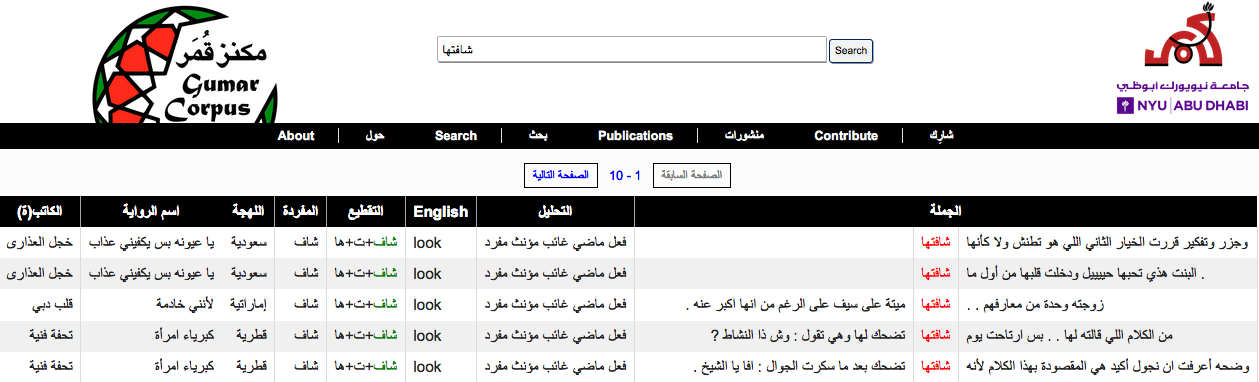}
\caption{Online web interface for browsing Gumar}\label{fig:website}
\end{center}

\end{figure*}

Following the collection of the corpus, we created a simple online interface that is specific for searching the corpus.\footnote{\url{http://camel.abudhabi.nyu.edu/gumar/}} The entire text of the corpus is stored in a relational database in an optimized manner. The lookup of the data is a simple search query that matches the user input to either a word token, lemma or stem form. Through the website interface, the rows of results are displayed to the user including the full context, word analysis that includes the POS, lemma, stem and gloss entries in addition to the information about the novel the word belongs to. See Figure~\ref{fig:website}.

\section{Conclusion and Future Work}
We collected the Gumar Corpus that consists of 100 million words from 1200 forum novels. We annotated the corpus for sub-dialect information at the document level of the novels in addition to the informations about the name and the writer's name of the novel. We also performed a preliminary investigation on the annotation of GA text. As an initial experiment we annotated around 4K words from four different novels using proposed orthography and morphology guidelines that followed previous efforts. We compared our gold annotations to the automatic annotations provided by MADAMIRA on its both MSA and EGY modes. The evaluation of the accuracy suggests that using MADAMIRA-EGY automatic annotations as a starting point for manual annotation of GA speeds up the process.

We plan to semi-automatically annotate the corpus and include a careful manual check at a large portion of it (1M words).  We are also looking forward to building a morphological analyzer for GA. We also plan to use the Gumar Corpus dialect annotations for some NLP tasks such as dialect identification.  We will make this corpus and its annotations publicly available.


\section{Acknowledgments}
We would like to thank Dimitrios Ntelitheos for helpful discussions. We would also like to thank Mustafa Jarrar and Rami Asia for sharing their set up of the Curras database browsing website.

%
\section{Bibliographical References}
\label{main:ref}
\bibliography{GulfArabicCorpus-LREC2016.bbl}


\end{document}